\pdfoutput=1

\documentclass[11pt]{article}

\usepackage{etoolbox}
\usepackage{listings}

\lstdefinelanguage{coq}{
  keywords={},
  keywordstyle=\textbf,
  basicstyle=\ttfamily,
  identifierstyle=\itshape\ttfamily,
  columns=fullflexible,
  keepspaces=true,
  mathescape=true,
  escapechar=!
}

\lstdefinestyle{normalstyle}{
  basicstyle=\fontsize{8}{8}\ttfamily,
  backgroundcolor=\color{yellow!10}}

\AtBeginEnvironment{lstlisting}{
  \lstset{style=normalstyle}}
  
\usepackage{EACL2023}
\usepackage{graphicx}
\usepackage{multicol}
\usepackage{multirow}
\usepackage{booktabs}

\usepackage{times}
\usepackage{latexsym}

\usepackage[T1]{fontenc}

\usepackage[utf8]{inputenc}

\usepackage{microtype}

\usepackage{inconsolata}

\title{Low-Resource Compositional Semantic Parsing with Concept Pretraining}

\author{Subendhu Rongali\thanks{~~work done when SR was at UMass and MS and KA were at Amazon; it does not relate to their current positions.} \\
  Amazon Alexa AI \\
  \texttt{rongals@amazon.com} \\\And
  Mukund Sridhar$^*$ \\
  Google \\
  \texttt{smukund@google.com} \\\And
  Haidar Khan \\
  Amazon Alexa AI \\
  \texttt{khhaida@amazon.com} \\\AND
  Konstantine Arkoudas$^*$ \\
  Dyania Health \\
  \texttt{konstantine@dyaniahealth.com} \\\And
  Wael Hamza \\
  Amazon Alexa AI \\
  \texttt{waelhamz@amazon.com} \\\And
  Andrew McCallum \\
  UMass Amherst \\
  \texttt{mccallum@iesl.cs.umass.edu} \\}

\newcommand{\ours}{\textsc{Seq2Seq-Ptr}}
\newcommand{\prop}{\textsc{Concept-Seq2Seq}}

\begin{document}

\maketitle

\begin{abstract}
Semantic parsing plays a key role in digital voice assistants such as Alexa, Siri, and Google Assistant by mapping natural language to structured meaning representations. To extend the capabilities of a voice assistant for a new domain, the underlying semantic parsing model needs to be retrained using thousands of annotated examples from the new domain, which is time-consuming and expensive. In this work, we present an architecture to perform such \emph{domain adaptation} automatically, with only a small amount of metadata about the new domain and without any new training data (zero-shot) or with very few examples (few-shot). We use a base seq2seq (sequence-to-sequence) architecture and augment it with a \emph{concept} encoder that encodes intent and slot tags from the new domain. We also introduce a novel decoder-focused approach to pretrain seq2seq models to be concept aware using Wikidata. This pretraining helps our model learn important concepts and perform well in low-resource settings. We report few-shot and zero-shot results for compositional semantic parsing on the TOPv2 dataset and show that our model outperforms prior approaches in few-shot settings for the TOPv2 and SNIPS datasets.
\end{abstract}

\begin{figure*}
    \centering
    \includegraphics[width=\textwidth]{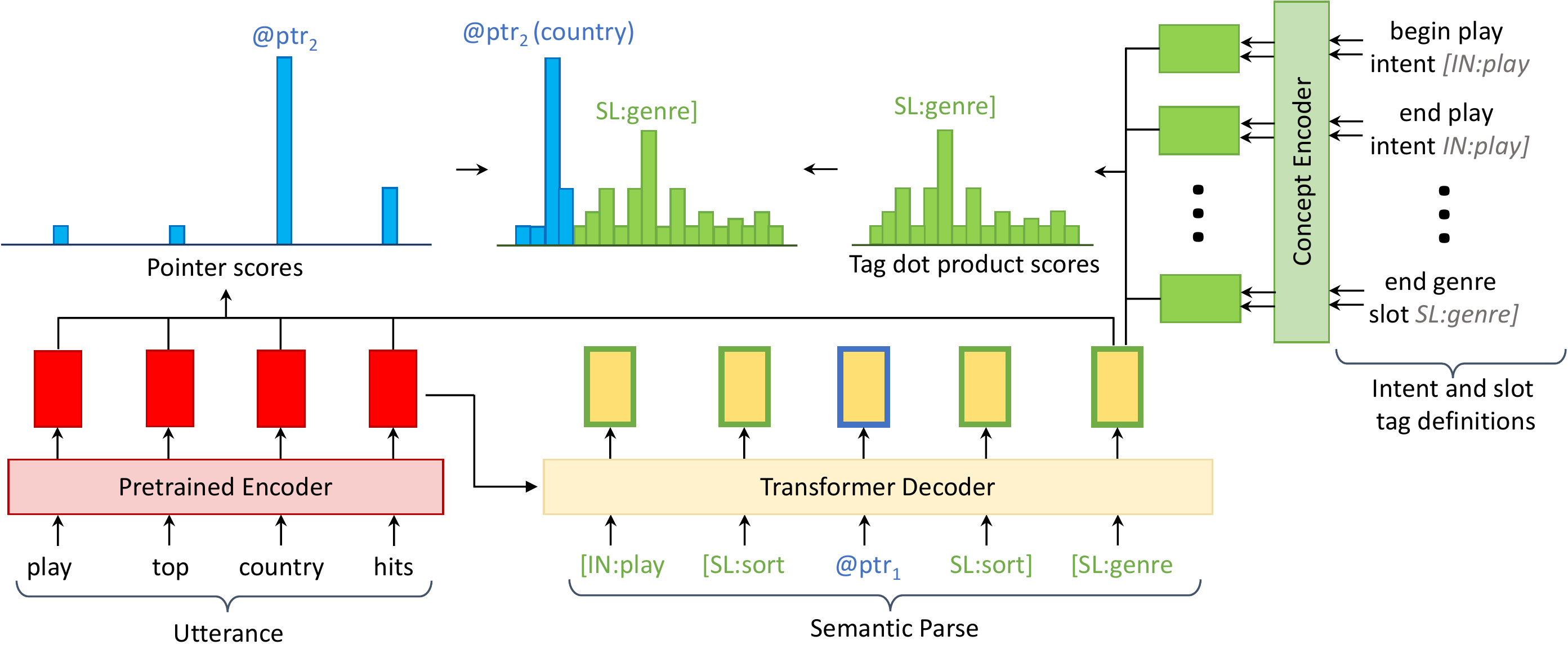}
    \caption{The architecture of {\prop} for low resource domain adaptation. The concept encoder encodes descriptions of each of the concept tags into an embedding and incorporates them into the decoded parse.}
    \label{fig:arch}
\end{figure*}

\section{Introduction}
Voice assistants such as Alexa, Siri, and Google Assistant often rely on semantic parsing to understand requests made by their users. The underlying semantic parsing model converts natural language user utterances into logical forms consisting of actions requested by the user (play music, check weather), called \emph{intents}, and relevant entities in the request (which song? which location?), called \emph{slots}. The model is built to process requests in a fixed set of domains, such as music, weather, shopping, and so on. With voice assistants increasingly pervading more aspects of daily life, systems need to be continuously updated to comprehend new intents and slots across an ever-growing number of domains.

Current semantic parsing models are trained on large amounts of annotated data from a predetermined set of domains. Extending these models to learn new intents or slots typically involves collecting and annotating large amounts of new data. This process is expensive and time-consuming. To combat this problem, researchers have proposed semantic parsing models that can be efficiently trained with fewer examples (few-shot) from new domains \citep{shrivastava2021span, mansimov2021semantic,ghoshal2020learning,shin2021constrained,desai2021low,rongali2022training,shrivastava2022retrieve}. While these methods facilitate few-shot learning, they have limitations. Some of them rely on hand-crafted knowledge such as intermediate grammars or logical form templates \citep{shrivastava2022retrieve,shin2021constrained,rongali2022training}. Others rely on very large pretrained language models, such as GPT-3, to perform in-context learning by appending test examples with instructional prompts \citep{shin2021constrained}.

In this work, we explore few-shot domain adaptation for semantic parsing without any additional hand-crafted knowledge apart from the intent and slot tag names, and with much smaller architectures that can perform efficient inference in practical production environments.
We also explore zero-shot domain adaptation, when we have no annotated training data from a new domain. 

To that end, we propose \prop, a novel architecture based on a state-of-the-art semantic parsing model, {\ours} \citep{rongali2020don}, which uses seq2seq models and a pointer generator network to decode the target semantic parse. We augment this model with a \emph{concept} encoder that encodes intents and slots from the schema and uses those encodings to conditionally decode the semantic parse. Figure~\ref{fig:arch} shows the architecture of our proposed model. We train this model on annotated data from the given domains. During inference, we simply encode all intents and slots from the schema, including new, unseen ones, into the learned concept space, and decode the target parse. This model has the same time complexity as the original {\ours} model but comes with the added benefit of now being able to effectively parse utterances from unseen domains without any additional effort. 

There have been a few zero-shot semantic parsing approaches proposed in the past but they either covered only simple slot-filling style utterances \citep{bapna2017towards,lee2019zero} or compositional utterances that also came with carefully crafted intermediate representations and context-free grammars \cite{herzig2018decoupling, wu-etal-2021-paraphrasing}. Our model is capable of performing zero-shot domain adaptation for compositional semantic parsing, producing meaning representations with nested intents and slots, but also doesn't require any grammars, whose construction effort often exceeds the effort required to annotate a few examples.

In few-shot scenarios, we fine-tune our zero-shot model checkpoints further on the small number of available examples. Due to the presence of the concept encoder in our architecture, we expect to receive better knowledge-transfer advantages by encoding intent and slot tags from new domains as opposed to initializing them as new tags. To further improve performance, we propose a novel decoder-focused pretraining scheme for {\prop} using an entity-centric processed version of Wikidata \citep{vrandevcic2014wikidata} called WikiWiki \citep{li22instilling}, to help it better encode unseen concepts and parse effectively.

We report the first zero-shot performance numbers for semantic parsing on the compositional TOPv2 dataset \cite{chen2020low} and show that {\prop} achieves commendable zero-shot performance on the flat-entity SNIPS dataset \cite{coucke2018snips}. We also evaluate in few-shot settings and show that we match or outperform previous state-of-the-art models while still being production-viable. 

In summary, our contributions are as follows.
\begin{itemize}
    \item We propose {\prop}, a bi-tower architecture with a seq2seq model and a concept encoder, that can perform few-shot and zero-shot domain adaptation for compositional semantic parsing without additional handcrafted knowledge.
    \item We propose a novel decoder-focused pretraining scheme for {\prop} using Wikidata that helps it better encode unseen concepts and parse effectively.
    \item We report few-shot and zero-shot semantic parsing results on the TOPv2 and SNIPS datasets and show that our model outperforms or matches previously proposed approaches on a variety of few-shot settings.
\end{itemize}

\section{Methodology}

In this section, we describe our proposed model, {\prop}, for low resource (few-shot and zero-shot) domain adaptation for semantic parsing. It is based on the {\ours} model from \citet{rongali2020don}, consisting of a sequence-to-sequence encoder-decoder component, augmented with a pointer generator network to constrain the target decoding vocabulary. Since our task at hand is to perform potential zero-shot semantic parsing with just descriptive metadata about the new domain, we modify the architecture of {\ours} to incorporate information about new intents and slots from new domains by adding a concept encoder. Section~\ref{arch} describes this architecture in detail. To help our model learn to parse utterances from unseen domains better, we also propose a novel pretraining scheme to incorporate general concept parsing knowledge into it. Section~\ref{cpret} describes this concept pretraining scheme. Finally, we describe {\prop} model specifics for few-shot and zero-shot settings in Section~\ref{specs}. Before we get to these sections, we first describe the source and target sequence formulation for the semantic parsing task below.

\subsection{Task Formulation}
Our model solves semantic parsing as a sequence-to-sequence task, where the source sequence is the utterance and the target sequence is a linearized representation of the semantic parse. Following \citet{rongali2020don}, we modify the target sequence to only contain intent/slot tags or pointers to utterance tokens. An example source and target sequence from the TOPv2 dataset are given below.

\begin{lstlisting}
Source: How far is the coffee shop
Target: [IN:GET_DISTANCE $@ptr_0$ $@ptr_1$ $@ptr_2$ 
  [SL:DESTINATION [IN:GET_RESTAURANT_LOCATION 
  $@ptr_3$ [SL:TYPE_FOOD $@ptr_4$ SL:TYPE_FOOD] 
  $@ptr_5$ IN:GET_RESTAURANT_LOCATION]
  SL:DESTINATION] IN:GET_DISTANCE]
\end{lstlisting}

Each $@ptr_i$ token here points to the $i^{th}$ token in the source sequence. So $@ptr_4$ corresponds to the word \emph{coffee}.

\subsection{Model Architecture}
\label{arch}

{\prop} consists of three main components: an encoder, a decoder, and a concept encoder. Just like in traditional sequence-to-sequence models, the encoder encodes the source sequence, and the decoder autoregressively decodes the target sequence. However, to effectively understand new intent and slot tags in target sequences that the model hasn't seen during training, our model needs to be able to incorporate new intents and slots, or concepts, and decode the target sequence accordingly. The concept encoder helps us do this by encoding descriptive metadata about new concepts and creating vector representations that we can use while decoding the target sequence.

Specifically, for an input sequence $[x_1 \dots x_n]$, we first encode it using the encoder into a sequence of hidden states $e_1 \dots e_n$. Then, having generated the first $t-1$ tokens, the decoder generates the token at step $t$ as follows. It first produces the decoder hidden state at time $t$, $d_t$ by building a multi-layer, multi-head self-attention on the encoder hidden states and the decoder states so far. This step is based on the transformer decoder from \citet{vaswani2017attention}. In a traditional sequence-to-sequence generation task, $d_t$ is then fed into a dense layer to produce scores over the target vocabulary. 

Our target vocabulary consists of pointer tokens and the concept tags. Since we do not have access to all concept tags at the time of training, we train our model to incorporate descriptive information about concepts instead of using a fixed-size dense layer. To do this, we encode intent and slot concepts using a \textbf{concept encoder}. The descriptions we use in this work are simply rule-based naturalized versions of the intent and slot names in the dataset. For example, the description for the intent tag token \texttt{[IN:GET\_DISTANCE} is set as ``\emph{begin get distance intent}''. Similarly, for \texttt{SL:DESTINATION]}, it is set to ``\emph{end destination slot}''. We purposely use just this information and no additional hand-crafted knowledge to remove any additional user input and to compare to previous approaches in the same setting. 

Given $m$ concept tokens (both begin and end) and their descriptions, the concept encoder encodes each of them to produce concept vector representations $[c_1 \dots c_m]$. We then use the computed decoder hidden state at $t$, $d_t$, as the query and compute unnormalized attention scores $[s_1 \dots s_m]$ with $[c_1 \dots c_m]$, and $[a_1 \dots a_n]$ with $[e_1 \dots e_n]$. Concatenating all these scores, we obtain an unnormalized distribution over $m + n$ tokens, the first $m$ of which are the intent and slot tagging tokens from the concepts, and the last $n$ of which are the $@ptr_{i} (0 < i < n)$ tokens pointing to the source sequence. We feed this through a softmax layer to obtain the final probability distribution. This probability distribution is used in the loss function during training and to choose the next token to generate during inference. For the target token embeddings in the decoder, we use a set of special embeddings to represent the $@ptr_i$ tokens and $[c_1 \dots c_m]$ to represent the concept embeddings.

Figure~\ref{fig:arch} shows this process in action on a toy example. The model is decoding the next token after \texttt{SL:genre} at step 5. To do this, the model computes the pointer attention scores $[a_1 \dots a_n]$ (blue, left) and the concept token attention scores $[s_1 \dots s_n]$ (green, right). The highest overall score is for the token $@ptr_2$, corresponding to the word \emph{country} in the source sequence, so the next predicted token is \emph{country}.

\begin{figure}
    \centering
    \includegraphics[width=0.4\textwidth]{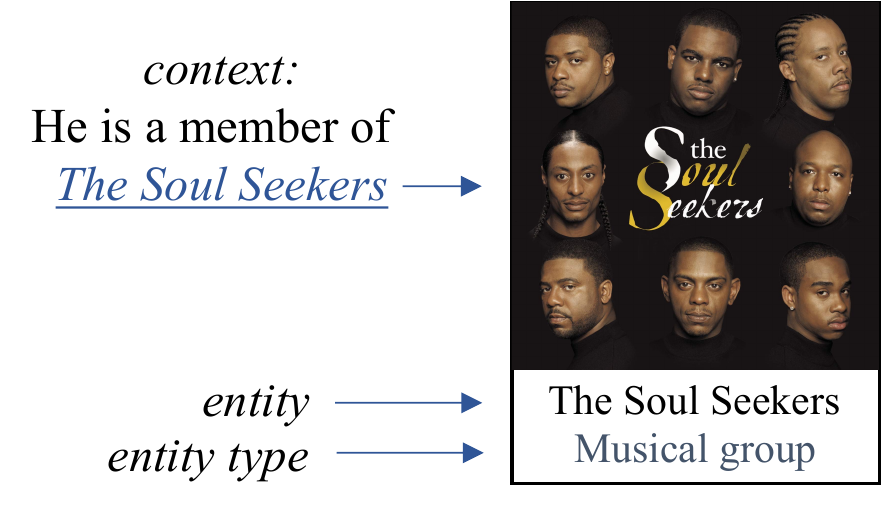}
    \caption{An example sentence from the Wikiwiki dataset with the associated mention, entity, and type fields. The full hyperlinked sub-span is extracted as the mention and the entity and type are extracted from the target page.}
    \label{fig:wikiwiki}
\end{figure}

\subsection{Concept Pretraining}
\label{cpret}

{\prop} has the ability to incorporate new, unseen concepts while parsing using the concept encoder and transfer knowledge on similar concepts. In order to produce these unseen concepts or types, and have our model be robust in low-data settings, it is important for our decoder to be type aware. Conventional seq2seq pretraining schemes such as \citet{lewis-etal-2020-bart,2020t5,atm20b} pretrain the decoder using language modeling criterion. \citet{li22instilling} extend the language modeling task to induce entity-type information by treating it as a question answering task. We pretrain the seq2seq model on the semantic parsing task using the WikiWiki \citep{li22instilling} dataset. We explain how to achieve this by keeping an open-domain, extensible output space for the semantic parse.

The Wikiwiki dataset curates mentions, entities, and entity types from 10M Wikipedia documents using hyperlink information linking sub-spans of text in sentences to other Wikipedia pages. The hyperlink is considered as the mention, and the entity and the type information are extracted from the new page. For further details on this processing, please refer to \citet{li22instilling}. This dataset contains around 2m entities and 40K entity types. 

Each example in the Wikiwiki dataset consists of a \emph{context}, which is a paragraph from a wiki page, \emph{mentions}, which are sub-spans of text that link to another page, \emph{entities}, which correspond to each mention, and \emph{entity types}, which describe the type of the entity. We extract individual sentences from this dataset and use them to train {\prop} to learn to encode a wide variety of concepts using the entity type fields as descriptions and tag the relevant mentions in the sentence. Figure~\ref{fig:wikiwiki} shows an example sentence from this dataset and the different fields. The source and target sequences for pretraining, and the descriptions of the concept tags for this example are given below.

\begin{lstlisting}
Source: He is a member of The Soul Seekers
Target: $@ptr_0$ $@ptr_1$ $@ptr_2$ $@ptr_3$ $@ptr_4$ 
   [Q215380 $@ptr_5$ $@ptr_6$ $@ptr_7$ Q215380]
Concept descriptions:
   [Q215380: begin musical group
   Q215380]: end musical group
\end{lstlisting}

During training, we collect all the concept tokens within a training batch and use them to create in-batch negatives (denominator of the softmax calculation) for the decoding task. We do this since it is extremely inefficient to encode all 40K $\times$ 2 concept token descriptions (begin and end) from Wikiwiki in every step.

\begin{table*}
\centering
\resizebox{\textwidth}{!}{ 
\begin{tabular}{lcccccccc}\toprule
& \bf Alarm & \bf Event & \bf Timer & \bf Weather & \bf Alarm & \bf Event & \bf Timer & \bf Weather \\\midrule
& \multicolumn{4}{c}{F1 score} & \multicolumn{4}{c}{EM Accuracy}\\
\midrule
{\prop} w/o pretraining &  62.53 & 30.25 & 55.68 & 51.23 & 45.94 & 0.00 & 0.51 & 8.62 \\
{\prop} & 71.00 & 70.47 & 58.48 & 66.29 & 53.64 & 20.21 & 3.86 & 26.63
\\\bottomrule
\end{tabular}}
\caption{Zero-shot performance of {\prop} on domains in TOPv2. We observe notable scores in the alarm and weather domains and improvements across all the domains after the concept pretraining step.}
\label{tab:ztop}
\end{table*}

\subsection{Few-shot and Zero-shot Specifics}
\label{specs}
{\prop} is primarily designed to perform low resource domain adaptation for semantic parsing by effectively encoding the output space via a concept encoder.

In the zero-shot setting, we deploy the following procedure to build our models. We first perform concept pretraining on {\prop} using Wikiwiki example sequences. We take this checkpoint and train on a set of known domains to then obtain the zero-shot model. During inference, we encode all the intent and slot tags from the new unknown domain using the concept encoder of the obtained zero-shot model and set the appropriate decoder parameters to reduce the architecture to a simple encoder-decoder setting.

In few-shot settings, we further fine-tune the zero-shot checkpoint on the available handful of training examples. Since we explore extremely low-resource settings (1, 5, 25 samples per intent/slot), we run the risk of over-fitting and instability during training. To account for these risks and smooth training, we augment the finetuning loss at every step with the loss from a randomly sampled batch of training data from the known domains. We scale the loss from the random known domain batch down using a multiplier before adding to the loss. This scheme is akin to rehearsal \citep{ratcliff1990connectionist}, a popular technique in domain adaptation.

\begin{table*}
\centering
\resizebox{\textwidth}{!}{ 
\begin{tabular}{lccccccc}\toprule
& \bf Music & \bf Book & \bf Creative & \bf Weather & \bf Restaurant & \bf Playlist & \bf Screening \\\midrule
\multicolumn{8}{c}{F1 score}\\
\midrule
Slot-filling \citep{lee2019zero,bapna2017towards} &  31.20 & \bf 34.13 & 86.21 & 65.64 & 51.40 & \bf 59.96 & 44.50 \\
{\prop} & \bf 50.00 & 30.26 & \bf 88.75 & \bf 74.58 & \bf 57.78 & 57.11 & \bf 45.78 \\
\midrule
\multicolumn{8}{c}{EM Accuracy}\\
\midrule
Slot-filling \citep{lee2019zero,bapna2017towards} &  11.00 & 1.00 & 69.00 & 37.00 & 12.00 & \bf 21.00 & 24.00 \\
{\prop} & \bf 20.00 & \bf 2.00 & \bf 69.00 & \bf 42.00 & \bf 13.00 & 19.00 & \bf 26.00 \\
\bottomrule
\end{tabular}}
\caption{Zero-shot performance of {\prop} on domains in SNIPS. Our model matches or outperforms the slot-filling style baseline on most domains.}
\label{tab:zsnips}
\end{table*}

\section{Experimental Setup}

We evaluate {\prop} on few-shot and zero-shot domain adaptation using two popular English task-oriented semantic parsing datasets - TOPv2 \citep{chen2020low} and SNIPS \citep{coucke2018snips}. Both datasets have utterances grouped into multiple domains; TOPv2 has eight domains and SNIPS has seven intents from seven different areas, which we consider domains. TOPv2 is a large dataset consisting of 10k-20k training and 3k-7k test examples per domain. It is also comprised of compositional examples with nested intents and slots. We exclude the \emph{unsupported} utterances from the training and test sets in TOPv2 for the zero-shot experiments (we use the full sets in few-shot). \emph{Unsupported} utterances consist of utterances that belong to a domain but are not supported, which is impossible to learn in zero-shot. SNIPS is a smaller and simpler dataset with flat, disjoint slots. It has 2k training and 100 test examples per domain.

For zero-shot, we use a leave-one-out approach where given $n$ domains, we train models on annotated data from $n-1$ of them and evaluate on utterances from the left-out domain. For few-shot settings, where the model has access to a few annotated examples from the left-out domain, we further fine-tune using 1, 5, and 25 samples-per-intent/slot (SPIs), which we randomly sample from the training data of the left-out domain. We fine-tune {\prop} on three different randomly sampled training sets per domain per SPI setting and report the average performance score of the three runs.

We use a transformer encoder, initialized from a \emph{roberta-base} checkpoint, for {\prop}. The decoder is a transformer decoder initialized from scratch and it contains 6 layers, 8 heads, and a hidden state size of 768. The concept encoder is also a transformer encoder and it is initialized from a \emph{bert-base-uncased} checkpoint. We choose a BERT-based model here since it is pretrained to compute a vector for the whole sequence using the CLS token, which is what we need for encoding a concept consisting of a multi-word description. We also choose all \emph{base}-size components to keep the overall model size small and expect the relative improvements shown by our model to generalize.

We train our zero-shot models using sequence cross entropy loss. We use the Adam optimizer with learning rate $2e^{-5}$ and $\epsilon = 1e^{-8}$, warm-up proportion $0.1$, weight decay $0.01$, and batch-size 128. The number of epochs is set to 100 and we evaluate after every epoch and early stop with a patience of 5. For the Wikiwiki pretraining step, we use the same hyper-parameters but stop the model training after 2 epochs on the entire Wikiwiki dataset. We did not perform explicit hyper-parameter tuning.

For the few-shot experiments, we take the zero-shot model trained by excluding the given few-shot domain and fine-tune it on small set of annotated examples for 1000 epochs, evaluating after every 25 epochs. All other parameters are set to the same values as in the initial zero-shot training. We use a multiplier of 0.1 for the augmented loss from the random known-domain data batch. To speed up evaluation during training, we use teacher-forced sequence accuracy as our validation metric, which doesn't require us to perform any beam search. During inference, we use beam search decoding with a beam size of 4.

We report exact match (EM) accuracy for the few-shot experiments, meaning the entire predicted parse has to match exactly with the gold parse. For zero-shot, we report both EM and F1 score since the task is more difficult and the performance is generally lower. At these lower numbers, F1 score, which awards partial credit to correctly tagged spans provides a better picture for improvements than EM accuracy, which requires the entire predicted parse to be correct for credit. For comparison wherever applicable, we use prior state-of-the-art models as baselines. In addition, we also report the performance of a vanilla {\ours} model without concept pretraining.

\section{Results and Discussion}

In this section, we report and discuss the performance of {\prop} on zero-shot and few-shot domain adaptation for semantic parsing. We first briefly describe our findings in zero-shot setting and then describe findings in a variety of few-shot scenarios.

\subsection{Zero-shot Domain Adaptation}
We report the first zero-shot performance numbers for domain adaptation on the TOPv2 dataset. Table~\ref{tab:ztop} contains these numbers. We observed that our model produced decent predictions on four of the eight domains in the dataset, which we document in the table. For the other domains, the scores were very low. {\prop} achieves good F1 and EM Accuracy scores on the alarm domain (71.00\% F1 and 53.64\% EM). On the weather and event domains, the concept pretraining step helps it achieve decent EM scores around 20\%. On the timer domain, {\prop} achieves a fairly high F1 score (58.48\%) but a very low EM score (3.86\%). Upon manual examination, we found that this was because our model always skipped a certain tag. In the timer domain, there is a slot tag called \texttt{SL:METHOD\_TIMER} which tags the kind of timer such as \emph{timer} or \emph{stopwatch}. Our model never learns to tag these words with that slot. We believe this is probably due to the description being inadequate for performing the requisite task. 

Overall, we believe the task at hand here is difficult due to the combination of the zero-shot setting and the presence of specific nesting/parsing rules in a compositional semantic parsing task. While a good amount of information can be gleaned from the intent and slot names, our model has no access to any new kinds of tagging rules since it has no annotated data or any descriptions of those rules within the concept descriptions. The descriptions themselves are sometimes inadequate as described with the timer domain above. We simply use the descriptions from the dataset and they weren't really designed to be used to describe the entity being tagged. We leave exploration into better descriptions and incorporating parsing rules without explicit annotations for future work.

We also evaluated our model on the SNIPS dataset to compare {\prop} to prior zero-shot approaches for flat slot-filling style datasets. We created a strong baseline using recent NLP advancements such as pretrained transformers and attention mechanisms on the slot-filling style zero-shot model proposed by \citet{bapna2017towards} and \citet{lee2019zero}. Table~\ref{tab:zsnips} compares the performance of {\prop} to this baseline. We observe that our model matches or outperforms the slot-filling baseline on most domains while also being adaptable to compositional datasets.

\begin{table*}
\centering
\resizebox{\textwidth}{!}{ 
\begin{tabular}{lcccccccc}\toprule
& \bf Alarm & \bf Event & \bf Messaging & \bf Music & \bf Navigation & \bf Reminder & \bf Timer & \bf Weather
\\\midrule
 \multicolumn{9}{c}{Few-shot 1 SPIs} \\\midrule
{\ours} & 20.41 & 31.85 & 38.12 & 25.58 & 19.96 & 23.66 & 16.62 & 47.24 \\
Inventory \cite{desai2021low} & 62.13 & 46.57 & \bf 46.54 & 23.00 & 21.16 & 28.58 & 28.92 & 54.53
 \\
RAF \cite{shrivastava2022retrieve} & 62.71 & - & - & 35.47 & - & - & \bf 55.06 & \bf 61.05 \\
{\prop} w/o pretraining & 61.72 & 44.28 & 34.24 & 20.66 & 20.82	& 35.39 & 44.75 & 52.24 \\
{\prop} & \bf 64.71 & \bf 54.42 & 46.13 & \bf 36.30 & \bf 30.00 & \bf 36.93 & 53.44 & 54.68\\
\midrule
 \multicolumn{9}{c}{Few-shot 5 SPIs} \\\midrule
{\ours} & 45.50 & 38.31 & 52.79 & 48.75 & 43.38 & 36.37 & 54.79 & 49.94 \\
Inventory \cite{desai2021low} & 71.81 & 58.87 & \bf 63.72 & \bf 53.59 & 42.59 & 48.88 & 55.54 & 65.09 \\
{\prop} w/o pretraining & 71.32 & 53.73 & 51.52 & 45.96 & 50.71 & 50.83 & 58.89 & 66.65\\
{\prop} & \bf 74.17 & \bf 61.72 & 61.20 & 51.24 & \bf 56.76 & \bf 54.36 & \bf 63.13 & \bf 68.54 \\
\midrule
 \multicolumn{9}{c}{Few-shot 25 SPIs} \\\midrule
{\ours} & - & - & - & - & - & 55.7 & - & 71.6 \\
RINE \cite{mansimov2021semantic} & - & - & - & - & - & \bf 68.71 & - & 74.53\\
{\prop} w/o pretraining & 78.16 & 68.21 & 75.28 & 65.54 & 67.67 &  67.92 & 70.72 & 74.30 \\
{\prop} & \bf 79.87 & \bf 72.96 & \bf 80.45 & \bf 67.91 & \bf 70.94 & 67.76 & \bf 72.41 & \bf 76.44\\
\midrule
\multicolumn{9}{c}{Reference - Fully trained} \\
\midrule
{\prop} & 88.07 & 83.23 & 93.11 & 79.47 & 81.63 & 79.57 & 77.33 & 90.73\\
\bottomrule
\end{tabular}}
\caption{EM Accuracy scores of various models in few-shot settings on TOPv2. We see that our model outperforms prior approaches on many domains and settings, most notably in the 1 SPIs setting.}
\label{tab:ftop}
\end{table*}

\subsection{Few-shot Domain Adaptation}
We evaluated {\prop} in few-shot settings of 1, 5, and 25 samples per intent/slot. Table~\ref{tab:ftop} reports the EM accuracy scores of {\prop} and other recent baselines on TOPv2. We also report the performance of a fully trained {\prop} model on all the training data for reference and to show that the architecture of {\prop} is competitive with other state-of-the-art methods in the full-resource setting.

We evaluated with a range of SPIs to allow for comparison with a wide range of models focused on both extremely low-resource (1, 5 SPIs) and medium low-resource (25 SPIs) settings. We report numbers for the baselines from their original papers, so they are missing for some domains.

As shown in the table, {\prop} outperforms a vanilla {\ours}, Inventory \citep{desai2021low}, and Retrieve-and-Fill (RAF) \cite{shrivastava2022retrieve} models on most domains in the 1 SPIs setting. RAF scores are very close and the approach outperforms our model on two domains but it uses additional hand-crafted information such as handmade descriptions and examples for intents and slots, as well as an intermediate scenario-bank to retrieve templates from. {\prop} simply works off of the existing information in the dataset. In the 5 SPIs setting, it again outperforms the vanilla {\ours} and Inventory models on most domains and on average. Inventory is a similar model to ours where the lexical information from intents and slots is used to help better transfer knowledge in the low-resource setting. However, this information is prepended to the input sequence and this might cause input size issues for large inventories. In the slightly higher resource setting of 25 SPIs, {\prop} beats a vanilla {\ours} model and matches the performance of RINE \cite{mansimov2021semantic}, reported on two domains. 

Across all the SPIs settings, we see that there is a noticeable drop in performance of {\prop} without the Wikiwiki concept pretraining. This shows the effectiveness of the pretraining step in helping the model generalize to unseen concepts and domains better.

To wrap up our evaluation, we also report the performance of {\prop} on SNIPS in the few-shot settings described above. Table~\ref{tab:fsnips} shows these numbers. We can see that we almost catch up to a fully trained model by training with just 25 SPIs in this dataset on most domains except music. We believe the music domain probably requires a lot of samples to effectively identify the diverse set of entities in the domain.

Overall, we find {\prop} to be a very promising approach which achieves high performance scores in low resource domain adaptation. It is capable of doing this in both compositional and flat semantic parsing settings, without any additional hand-crafted information apart from the little documentation in the dataset, and with the memory and inference latency footprint of a vanilla {\ours} model. 

\begin{table*}
\centering
\resizebox{\textwidth}{!}{ \begin{tabular}{lcccccccccccccc}
\toprule
& \multicolumn{2}{c}{\bf Music} & \multicolumn{2}{c}{\bf Book} & \multicolumn{2}{c}{\bf Creative} & \multicolumn{2}{c}{\bf Weather} & \multicolumn{2}{c}{\bf Restaurant} & \multicolumn{2}{c}{\bf Playlist} & \multicolumn{2}{c}{\bf Screening} \\\midrule
& F1 & EM & F1 & EM & F1 & EM & F1 & EM & F1 & EM & F1 & EM & F1 & EM\\
\midrule
1 SPIs & 67.10 & 39.00 & 86.74 & 61.67 & 92.39 & 80.33 & 88.14 & 71.00 & 89.64 & 72.00 & 78.31 & 47.00 & 79.19 & 61.00 \\
5 SPIs & 84.33 & 68.00 & 96.83 & 89.33 & 92.58 & 84.33 & 94.64 & 86.33 & 93.27 & 81.33 & 87.74 & 68.33 & 94.16 & 87.67 \\
25 SPIs & 85.90 & 72.67 & 99.32 & 97.33 & 98.10 & 95.33 & 98.27 & 94.67 & 96.38 & 89.33 & 91.73 & 80.67 & 95.02 & 89.33\\
Fully-trained & 90.78 & 83.00 & 99.18 & 97.00 & 100.00 & 100.00 & 98.55 & 96.00 & 96.89 & 90.00 & 94.53 & 87.00 & 98.35 & 97.00\\
\bottomrule
\end{tabular}}
\caption{Few-shot performance of {\prop} on domains in SNIPS. Our 25 SPIs model almost catches up to a fully trained model. Numbers are an average of three runs with different random samples of SPIs.}
\label{tab:fsnips}
\end{table*}

\section{Related Work}
Zero-shot domain adaptation for task-oriented semantic parsing has been previously explored for simple flat queries with single intents and disjoint, non-overlapping slots. \citet{bapna2017towards} and \citet{lee2019zero} encode the lexical tag features and create a token-tagging schema to create the final semantic parses. \citet{yu2021few} solve the task using a retrieve-and-fill mechanism. Our baseline model for simple queries is based on these approaches.

For complex utterances with nested structures, zero-shot semantic parsing has been explored using intermediate, concept-agnostic logical forms \citep{herzig2018decoupling,dong2018coarse,reddy2017universal} or natural language canonical forms \citep{wu-etal-2021-paraphrasing}. These approaches apply to semantic parsing datasets which have context free grammars and specified rules, such as database or knowledge graph queries. The effort to craft these grammars for task-oriented semantic parsing in a voice assistant setting could quite possibly be greater than annotating utterances.

A more relevant class of approaches for this work are ones that solve task-oriented semantic parsing for complex utterances in a few-shot setting using lexical tag features. \citet{shrivastava2021span} and \citet{mansimov2021semantic} modify the seq2seq architecture from \citet{rongali2020don} to perform non-autoregressive style decoding and show that their models perform better in a few-shot setting. \citet{ghoshal2020learning} use adaptive label smoothing, a model-agnostic technique. \citet{shin2021constrained} proposed a prompting-style approach where custom instructional prompts filled with handful of annotated examples and an unsolved utterance are fed as input to GPT-3 to directly produce a semantic parse. Their approach is extremely slow and cannot be easily adapted into a zero-shot framework.
\citet{shrivastava2022retrieve} explore a retrieve-and-fill style approach where they retrieve the best \emph{scenario}, an intermediate logical form consisting of the semantic frame and abstracted out tags, from a scenario bank of all supported semantic parses. Their approach is contingent on the availability of this scenario bank which could possibly entail more effort than annotating utterances. \citet{mueller2022label} and \citet{desai2021low} use lexical features from intent and slot names to create an \emph{inventory} and use it as input to train semantic parsers for new domains. \citet{mueller2022label} also pretrain their model to improve generalizability but only evaluate it on an intent classification task. \citet{desai2021low} evaluate their model for full sequences and our model is similar to theirs. However, we use our inventory to create custom decoder embeddings in a seq2seq model, which removes any input size issues that their model will encounter with large inventories. We also pretrain our model with Wikidata and evaluate it in a completely zero-shot setting, in addition to few-shot. \citet{zhao2022compositional} is another recent question-answering-based approach that uses lexical features from the intent and slot tags by using them as context and posing questions but it has a similar input size issue with large inventories.

\section{Conclusion}

We propose a model called {\prop} to perform low-resource domain adaptation for compositional semantic parsing. Our model is built on the {\ours} framework and is augmented with a concept encoder to transfer knowledge and encode unseen intents and slots from new domains through their text definitions. We also propose a novel concept pretraining scheme to incorporate general concept knowledge into our model using an entity-centric Wikipedia dataset called Wikiwiki.

We evaluate our model in zero-shot and multiple few-shot settings on Facebook TOPv2 and SNIPS datasets. We show that our model is capable of performing zero-shot domain adaptation on some domains of the TOPv2 dataset and beats a strong slot-filling baseline on the SNIPS dataset. In few-shot, over multiple dataset sizes of 1, 5, and 25 SPIs, we show that our model outperforms many strong prior models on TOPv2. Using the SNIPS dataset, we also demonstrate how our model catches up to a fully-trained semantic parsing model using just 25 SPIs on most domains. Our model is capable of low-resource domain adaptation in both compositional and flat parsing settings, without additional hand-crafted information, and with the inference behavior of a vanilla {\ours} model.

\section*{Acknowledgements}
This work was supported in part by Amazon Alexa AI, in part by the Chan Zuckerberg Initiative, in part by IBM Research AI through the AI Horizons Network, and in part by the National Science Foundation (NSF) grants IIS-1763618 and IIS-1955567. Any opinions, findings, conclusions, or recommendations expressed in this material are those of the authors and do not necessarily reflect those of the sponsors.

\section*{Limitations}
Our models were trained on GPUs that had at least 20GB on-board memory since in addition to the traditional encoder and decoder components in a seq2seq model, we also train a concept encoder, which is around the same size as the encoder. This eliminates the use of popular GPUs such as 1080-ti and 2080-ti, unless parameter freezing or other tricks are employed during training. During inference however, once the concepts are trained, we can simply encode the target tags and this reduces to the size and performance of a traditional {\ours} model.

We also report all results on models (ours and baselines) with \emph{base}-size components such as \emph{roberta-base}. We do this since these models are more likely to be used in production than the much bigger \emph{large}-size models. Results and comparison with \emph{large}-sized models is missing from this work (we expect the trends shown to generalize) and we leave this to future work.

Finally, to simulate our low-resource experiments, we randomly sample a few examples from the existing training datasets. While this is useful for experimentation, it doesn't truly mimic a real low-resource workflow where these few examples could be carefully crafted by developers to ensure better semantic coverage in terms of the language of the utterances. This work doesn't include any analysis on the influence of the content of the few selected examples; it just focuses on their number.

\bibliography{anthology,custom}
\bibliographystyle{acl_natbib}

\end{document}